\documentclass[9pt,journal,compsoc]{IEEEtran}

\ifCLASSOPTIONcompsoc
  \usepackage[nocompress]{cite}
\else
  \usepackage{cite}
\fi
\usepackage{url}
\usepackage{svg, todonotes, boldline, bbm, makecell, multirow, hyperref}
\usepackage{amsmath}
\usepackage{amssymb}
\usepackage{siunitx}

\ifCLASSINFOpdf
\else
\fi

\begin{document}

\title{Learning Probabalistic Graph Neural Networks for Multivariate Time Series Anomaly Detection}

\author{
       Saswati~Ray$^*$
       Sana~Lakdawala$^*$,
       Mononito~Goswami$^*$,
       Chufan~Gao$^*$,
       \protect\\
       \{saswatir, slakdawa, mgoswami, chufang\}@andrew.cmu.edu
\thanks{\textbf{$^*$ denotes equal contribution}}
}

\markboth{Learning Graph Neural Networks for Multivariate Time Series Anomaly Detection}%
{Gao \MakeLowercase{\textit{et al.}}: Final Summary: Learning Graph Neural Networks for Multivariate Time Series Anomaly Detection}

\IEEEtitleabstractindextext{%

}
\maketitle
\IEEEdisplaynontitleabstractindextext
\IEEEpeerreviewmaketitle

\ifCLASSOPTIONcompsoc
\IEEEraisesectionheading{\section{Introduction}\label{sec:introduction}}
\else
\section{Introduction}
\label{sec:introduction}
\fi

\IEEEPARstart{A}{n} anomaly is loosely defined as any observation which deviates so much from the remaining observations, so as to arouse suspicions that it was generated by a different mechanism \cite{hawkins1980identification}. In most applications, data is generated by a \textit{data generating process}. Anomalies are generated when this process behaves unusually. Therefore, anomaly detection can provide several useful application specific insights, \textit{e.g.} from detecting credit card fraud to tracking interesting sensor events. 
Consequently, detecting anomalies in high-dimensional multivariate time series (MTS) is an extremely important problem which has been extensively studied in literature \cite{aggarwal2015outlier}. However, most prior work on anomaly detection does not explicitly  model the complex dependencies between different variables, which significantly limits their ability to detect anomalous events \cite{deng2021graph}. For instance, in a water treatment plant, if a motorized valve is maliciously turned on, it can cause an overflow on a tank, resulting in anomalous readings in most sensors associated with it\footnote{In the \texttt{WADI} dataset \cite{ahmed2017wadi}, the first attack involves maliciously turning on motorized valve \texttt{1\_MV\_001} which causes an overflow in the primary tank. This reflects in sensors \texttt{1\_LT\_001} and \texttt{1\_FIT\_001} associated with the primary tank.}. 
Recently, Deng et al. \cite{deng2021graph} proposed the Graph Deviation Network (GDN) which automatically learns variable dependencies and uses them to identify anomalous behaviour. Like most Neural Networks (NNs), despite having an impressive accuracy, GDN produces poor uncertainty estimates. Since overconfident yet incorrect predictions may be harmful, precise uncertainty quantification is integral for practical applications of such networks \cite{lakshminarayanan2016simple}. 
To this end, we propose \textbf{GLUE} (\textbf{G}DN with \textbf{L}ocal \textbf{U}ncertainty \textbf{E}stimation) which not only automatically learns complex dependencies between different variables in a MTS and uses them to detect anomalous behaviour, but also models the uncertainty of predictions. 

Results on two real world datasets reveal that GLUE performs on par with GDN, outperforms most popular baseline models, and also learns meaningful dependencies between variables (Sec.~\ref{experiments&results}). 
The rest of the paper is organized as follows. In Sec.~\ref{relatedwork} we briefly compare our work with prior work. Next, in Sec.~\ref{approach} we describe the GDN and GLUE models in detail followed by an overview of our baselines. Sec.~\ref{dataset} and~\ref{experiments&results} discuss our datasets, experimental setup and results. We conclude the paper with Sec.~\ref{conclusion}. 
\subsubsection{Primary Contributions}
Our primary contributions are as follows:
\begin{enumerate}
    \item \textbf{Uncertainty estimation}: GDN forecasts sensor readings as point estimates and like most NNs is likely to produce poor uncertainty predictions. In order to produce proper uncertainty estimates, GLUE instead predicts the mean $\mu(x)$ and variance $\sigma^2(x)$ of a heteroscedastic Gaussian distribution and treats the observations as a sample from this distribution. As against GDN which minimizes the Mean Squared Error (MSE) on the training set, GLUE optimizes Gaussian negative log likelihood \cite{nix1994estimating} of the observations, allowing us to address prediction uncertainty and ensure interpretable thresholds \cite{an2015variational, park2018multimodal, lakshminarayanan2016simple}.
    \item \textbf{Validation on a new dataset}: We evaluate our proposed approach against a suite of increasingly complex and popular baselines on two real world datasets, the \texttt{WADI} dataset from the original paper and a new \texttt{NASA Turbofan} dataset (Sec.~\ref{dataset})
    \item \textbf{Open Source Implementation}: We also release the code of our proposed GLUE model to ensure reproducibility (\href{https://github.com/chufangao/GLUE}{github.com/chufangao/GLUE}).
\end{enumerate}
\section{Related Work}
\label{relatedwork}
\subsection{Anomaly Detection}
There has been much work on time series anomaly detection. Angiulli and Pizzuti \cite{angiulli2002fast} proposed an unsupervised distance-based approach for outlier detection using sum of distances from a point's $k$-nearest neighbors. Later, Shyu et al. \cite{shyu2003novel} used an unsupervised linear-model approach for anomaly detection using Principal Component Classifiers. More recently, Aggarwal \cite{aggarwal2015outlier} discussed the use of the deep autoencoder-based reconstruction error as a measure of anomalousness. While these methods are popular and yield competitive results across a multitude of problems, they \textit{do not account for the complex dependencies between variables.}
Recently, Graph Neural Networks (GNNs) have been widely used to model structured data. For instance, Kipf and Welling \cite{kipf2016semi} used Graph Convolution Networks (GCNs) to model complex patterns for classification in graph-structured data, whereas Ye et al. \cite{yu2017spatio} approached the time series forecasting problem in traffic domain using Spatio-Temporal GCNs to capture comprehensive spatio-temporal correlations. In general, GNNs assume that the state of a node is influenced by the states of its neighbour nodes \cite{deng2021graph}. Both GLUE and GDN \textit{inherit this fundamental property of GNNs}. However, in contrast to GNNs, GLUE and GDN \textit{do not share the same set of parameters for each node}, which renders GNNs ineffective in modeling nodes with different behaviour. Moreover, \textit{GNNs require a graph structure} as input, whereas \textit{GLUE learns the graph structure from the data}. 
\subsection{Uncertainty Estimation}
Quantifying uncertainty in NNs predictions is challenging and an as yet unsolved problem \cite{lakshminarayanan2016simple}. Hence, GDN \cite{deng2021graph} like most NNs is likely to produce overconfident uncertainty estimates \footnote{We must note that GDN technically does does not provide any uncertainty estimates and just forecasts point estimates on its forecasted values. Hence, our argument is based on the fact that NNs in general produce extremely poor uncertainty estimates.}. In contrast, GLUE predicts the mean and variance of a heteroscedastic multivariate Gaussian distribution and treats the observed sensor readings as samples from this distribution, following the seminal work by Nix and Weigend \cite{nix1994estimating}.    
\section{Approach}
\label{approach}
\begin{figure}[!ht]
  \centering
   \includegraphics[width = 0.5\textwidth]{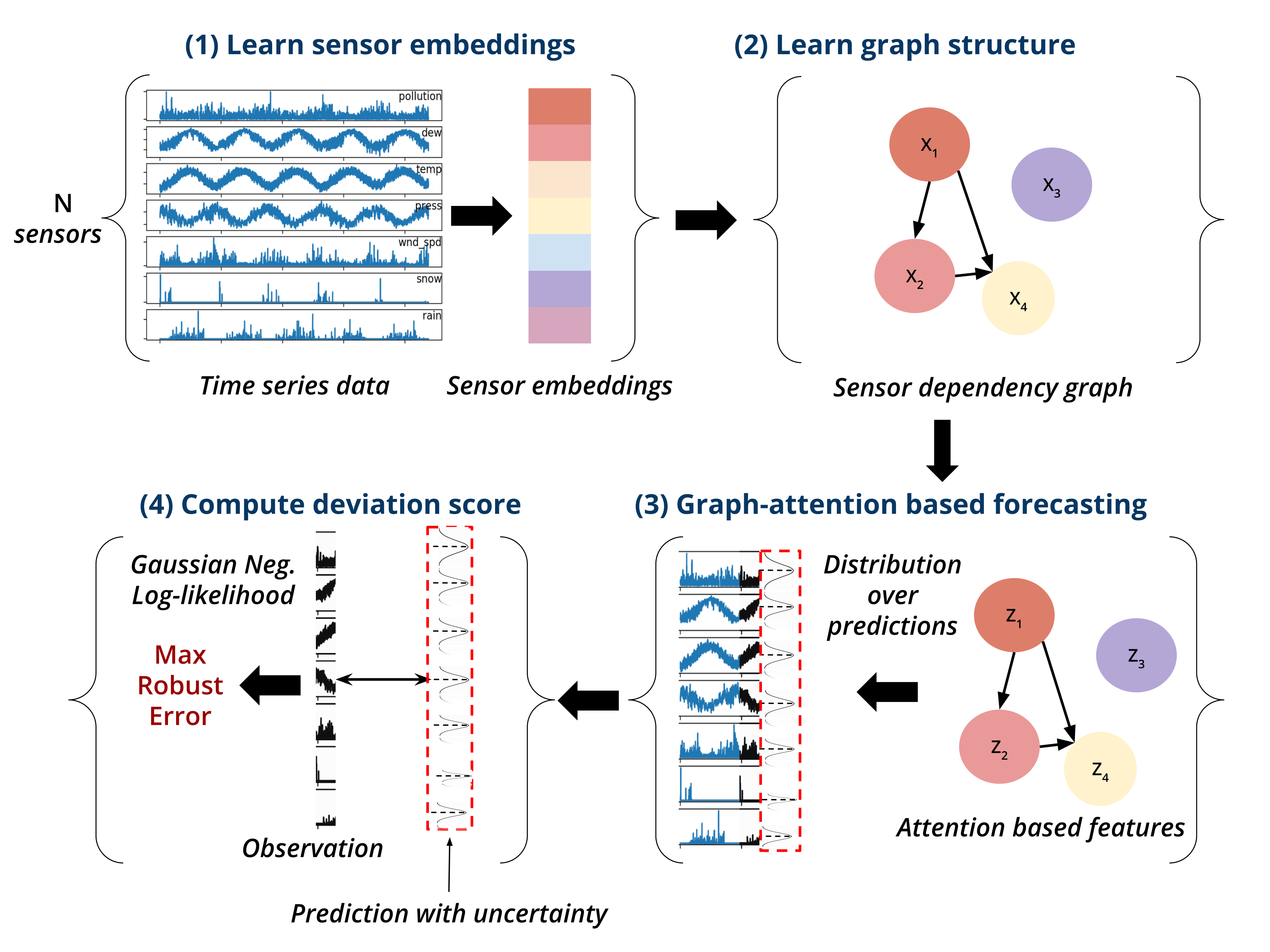}
  \caption{\textbf{Graph Deviation Network with Local Uncertainty Uncertainty Estimation (GLUE)} learns complex dependencies between variables of a MTS and uses it to identify anomalous behaviour.}
  \label{fig:graphdeviationnetwork}
\end{figure}
Our primary goal is to learn the dependencies between the $N$ sensors of a system (\emph{structure learning}) and use them to learn a model $\mathcal{M}$ of normal system operation. We then use $\mathcal{M}$ to flag significant deviations from the expected sensor values as anomalies (\emph{structured prediction}). The key ingredients of \textbf{GLUE} are as follows--
\\\\
\textit{1. Learn sensor embeddings}: Sensor embeddings $\textbf{V} = \{\textbf{v}_1, \dots, \textbf{v}_N\}$,  capture the unique characteristics of each of the $N$ sensors. Here, each $\textbf{v}_i \in \mathbb{R}^d, \forall i \in \{1\dots N\}$, where $d$ corresponds to the dimensionality of the embedding vector. 
\\\\
\textit{2. Learn sensor dependencies}: Next, we learn the relationships between sensors using their embeddings and encode them as edges into a directed graph $\mathcal{G}$ (neighbour graph). This is done by learning a sparse adjacency matrix $\mathcal{A}$ using top $k$ neighbors for each sensor $i$. The parameter $k$ controls the sparsity of the resulting neighbour graph $\mathcal{G}$. 
  \begin{align}
      e_{ji} &= \dfrac{\textbf{v}_i^T\textbf{v}_j}{\|\textbf{v}_i\|\cdot\|\textbf{v}_j\|}, \ \forall j \in \{1, \dots, N\} \\
      \mathcal{A}_{ji} &= \mathbbm{1}\{j \in \texttt{TopK}(\{e_{ki} : k \in \mathcal{C}_i\})\}
  \end{align}
  where $\mathcal{C}_i \subseteq \{1, \dots, N\} \setminus \{i\}$. $e_{ji}$ is simply the cosine similarity between the embeddings of any two sensors $i$ and $j$, whereas \texttt{TopK} returns the indices of the $k$ nearest nodes. $\mathcal{C}_i$ refers to the number of candidate relationships for sensor $i$. Domain experts can specify sensors which are more likely to be related to each other. However, if we are not aware of any relationships apriori, we can set $\mathcal{C}_i = \{1, \dots, N\}$.   
\\\\
\textit{3. Predict sensor readings at time $t$} Given a fixed window size $w > 0$ of historical sensor readings from time steps $(t - w)$ to $(t - 1)$, we predict $\mathbf{\hat s}_i^{(t)} \sim \mathcal{N}(\mu_\theta(\mathbf{x}_i^{(t)}), \sigma_\theta^2(\mathbf{x}_i^{(t)}))$ \textit{i.e.} the heteroscedastic Gaussian distribution around the future sensor reading for $i$'th sensor at time $t$ based on an attention function over its top-$k$ neighbors $\mathcal{N}(i)$ in $\mathcal{A}$. Here, $\theta$ represent the GLUE parameters. In GLUE, as in GDN for each sensor $i \in \{1, \dots, N\}$, we subject its historical readings, $\mathbf{x}_i^{(t)} =  [\mathbf{s}_i^{(t-w)},\dots,\mathbf{s}_i^{(t-1)}], \ \mathbf{x}_i^{(t)} \in \mathbb{R}^{w}$ to the following transformations. Let $\textbf{W} \in \mathbb{R}^{d\times w}$ and $\mathbf{a}_i\in \mathbb{R}^{4d\times 1}$ denote a set of weights and attention mechanism coefficients, respectively.
Instead of forecasting the next $\mathbf{\hat s}_i^{(t)}$ as GDN does, GLUE predicts the mean ($\mu_\theta(\mathbf{x}_i^{(t)})$) and variance ($\sigma_\theta^2(\mathbf{x}_i^{(t)})$) of a gaussian distribution and treats the future reading as a sample from this distribution. We compute $\mu_\theta(\mathbf{x}_i^{(t)})$ and $\sigma_\theta^2(\mathbf{x}_i^{(t)})$ as follows--
  \begin{align}
      \mathbf{g}_i^{(t)} &= \textbf{v}_i \oplus \textbf{W}\mathbf{x}_i^{(t)}      \\
      \pi(i,j) &= \texttt{LeakyReLU}(\mathbf{a}^T(\mathbf{g}_i^{(t)} \oplus \mathbf{g}_j^{(t)})), \ j\in \mathcal{N}(i) \\
      \alpha_{i,j} &= \dfrac{\exp(\pi(i,j))}{\sum_{k\in N(i)\cup \{i\}}\exp(\pi(i,k))} \\
      \mathbf{z}_i^{(t)} &= \texttt{ReLU}\left(\alpha_{i,i}\textbf{W}\mathbf{x}_i^{(t)} + \sum_{j\in \mathcal{N}(i)}\alpha_{i,j}\textbf{W}\mathbf{x}_j^{(t)}\right) \\
      \mu_\theta(\mathbf{x}_i^{(t)}), \ & \sigma^2_\theta(\mathbf{x}_i^{(t)}) = f_{\phi}(\textbf{v}_i\odot \mathbf{z}_i^{(t)})
     \\
     \hat{s}_i^{(t)} & := \mu_\theta(\mathbf{x}_i^{(t)})
  \end{align}
  where, $f_{\phi}$ are stacked fully-connected layers. The sensor prediction for the $t^{th}$ timestep ($\hat{s}_i^{(t)}$) is defined as the Maximum Likelihood Estimate (MLE) of the predicted Gaussian distribution \textit{i.e.} its mean. 
\\\\
\textit{4. Identify deviations} Finally, we identify anomalies as significant deviations of the ground truth from the expected behaviour (explained in Baselines and Metrics). In contrast to vanilla GDN which predicts point estimates $\mathbf{\hat s}^{(t)}$, GLUE \textit{will} produce distributions $\mathbf{\hat s}^{(t)}$ with uncertainty estimates. 

\subsection{Baselines and Metrics}
We compare GLUE with a host of popular baseline models: Principal Components Analysis (PCA) \cite{shyu2003novel}, $k$-nearest neighbors (KNN) \cite{angiulli2002fast}, and autoencoders (AE) \cite{aggarwal2015outlier}, which do not consider the complex structural relationships between the sensors. Since GDN and GLUE essentially forecast the next sensor readings, we also compare them with an advanced forecasting model Vector Autoregression (VAR) \cite{ford1986beginner}.  

\begin{figure}[!hbtp]
    \centering
     \begin{tabular}{cc}
	 \includegraphics[width=0.23\textwidth]{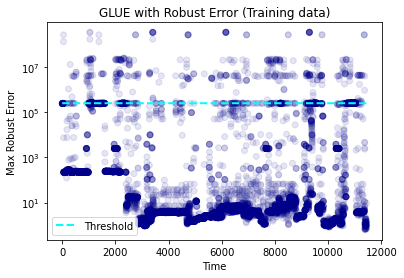} &
    \includegraphics[width=0.23\textwidth]{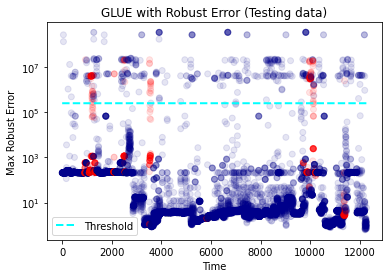} \\
    (a) WADI train & (b) WADI test \\
    \includegraphics[width=0.23\textwidth]{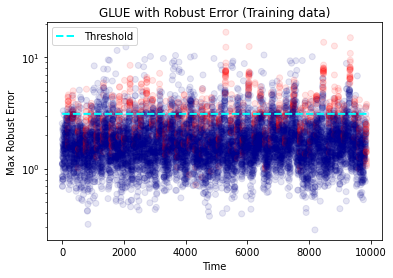} &
    \includegraphics[width=0.23\textwidth]{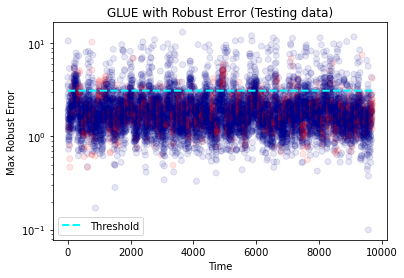} \\
    (a) NASA train & (b) NASA test \\
	 \end{tabular}
	 \caption{Plots showing the max robust error (GLUE) across all sensors over time for the WADI (top) and NASA (bottom) train (left)-test (right) datasets. The \textcolor{red}{red} points show the true anomalies in the respective datasets (none in WADI training data), whereas the \textcolor{blue}{blue} points represent the normal points for respective datasets. Cyan lines show the threshold (chosen as the proportion of expected anomalies in the training dataset) above which data points have been labeled as anomalies. }
    \label{fig:tuning}
\end{figure}

We ran each of the baselines on the same time window as input. We classified all test data points with a reconstruction error higher than the $\left(1-\frac{\# \text{num\_anomalies}}{\text{total\_points}}\right)^{th}$ quantile reconstruction error in the training data as anomalous.
For GDN and GLUE, we predicted anomalies on a robust standardized forecasting error referred to as the Max Robust Error (MRE). The MRE at the $t^{th}$ timestep is defined as the maximum robust forecasting error among all sensors:  
\begin{align}
    \text{MRE}(t) &= \max_{i \in \{1, \dots, N\}} \frac{|s_i^{(t)} - \hat{s_i}^{(t)}| - \tilde \mu_i}{\tilde \sigma_i}
\end{align}
where $|s_i^{(t)} - \hat{s_i}^{(t)}|$ is the absolute value of the difference between the observed and forecasted values for sensor $i$\footnote{Note that while GDN forecasts the sensor value, the forecast for GLUE instead is the MLE of the predicted Gaussian distribution.} and $\tilde \mu_i$ and $\tilde \sigma_i$ are the median and the interquartile range for a particular sensor $i$. Data points with MREs above a threshold (determined from the training data) were classified as anomalous (see Fig.~\ref{fig:tuning}).

\subsection{Training objective of GDN and GLUE}
While GDN minimizes the mean squared error between the observed sensor readings and the forecasted sensor values, \textit{i.e.} $(s_i^{(t)} - \hat{s_i}^{(t)})^2$, GLUE optimizes the negative Gaussian log likelhood \cite{nix1994estimating} of the observed readings based on its predicted mean ($\mu_\theta(\mathbf{x}_i^{(t)})$) and variance ($\sigma^2_\theta(\mathbf{x}_i^{(t)})$): 
\begin{align*}
    loss &\propto \sum_i^{N} \frac{\log (\sigma_\theta^2(x_i))}{2} + \frac{(y - \mu_\theta(x_i))^2}{2 \sigma_\theta(x_i)^2} 
\end{align*}
where $\mu_{\theta}, \sigma_{\theta}$ are output by the GLUE model and $\theta$ represent the GLUE model parameters. 
\section{Datasets}
\label{dataset}
Since large-scale real world datasets with labeled anomalies are scarce, we used two datasets which realistically simulate both normal and anomalous scenarios. For our experiments, we will consider the following two datasets.
In our experiments, both datasets were normalized for zero mean and a standard deviation of 1. Additionally, NaNs were first forward-filled, then back-filled, and then filled with zeroes to ensure stability and accuracy. To obtain the time window, we take 5 time steps of each sensor as input.
\subsection{\texttt{WADI} Dataset}
The \texttt{WADI} dataset curated by the Singapore Public Utility board comes from a realistic water treatment, storage and distribution network \cite{ahmed2017wadi}. 
The data comprises of $127$ sensor measurements at $1$Hz over $16$ days of continuous operation: $14$ under normal operation and $2$ days with abnormal attack scenarios resulting in total of $104,847$ training and $17,270$ testing data points with an anomaly rate of $5.99\%$. In consonance with the original paper, our results are reported for a downsampled version of the \texttt{WADI} dataset with a rolling median of 10 seconds over the entire time period. Additionally, we also filter out $30$ sensors with $0$ variance over the entire course of the 2 week period.
\subsection{NASA AMES' Turbofan Dataset}
The NASA Turbofan dataset \footnote{\href{https://data.nasa.gov/dataset/Turbofan-engine-degradation-simulation-data-set/vrks-gjie}{https://data.nasa.gov/dataset/Turbofan-engine-degradation-simulation-data-set/vrks-gjie}}\cite{chao2021aircraft} is derived from an engine turbofan run under various climatic conditions until it failed. 
This dataset consists of numerous trajectories of multiple time steps each (a total of $157,523$ train points and $102,069$ test points, with about $10$\% of the data classified as anomalies, or, engine failures), for which readings of $24$ sensors are recorded. One of the challenges this dataset poses is generalization over the climate conditions, as well as over degradation of various parts of the turbofan. In this dataset, there are two distinct failure modes, however our experiments will draw no distinction between the two, and both modes will be classified as the anomalies to be detected.


\begin{table}[!th]
  \centering
  \begin{tabular}{ccccc}
    \Xhline{1pt}
     \textbf{Datasets} & \textbf{\#Sensors} & \textbf{\#Train} & \textbf{\#Test} & \textbf{\%Anomalies}\\
     \Xhline{1pt}
     \texttt{WADI} & 127 & 104,847 & 17,270 & 6.89 \\
     \texttt{NASA Turbofan} & 24 & 157,523 & 102,069 & 10.82\\
     \Xhline{1pt}
  \end{tabular}
\vspace{1pt}
  \caption{Statistics of datasets used for our experiments.}
  \label{tab:data}
\end{table}

\section{Experiments \& Results}
\label{experiments&results}
\begin{figure*}[!htb]
    \centering
     \begin{tabular}{cccc}
	 \includegraphics[width=0.23\textwidth]{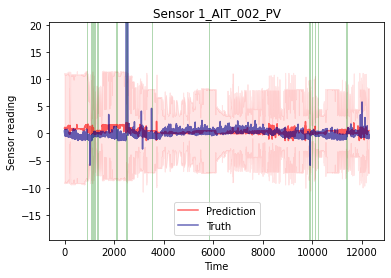} &
	 \includegraphics[width=0.23\textwidth]{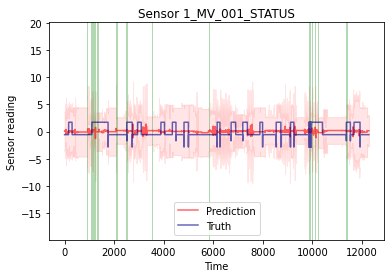} &
     \includegraphics[width=0.23\textwidth]{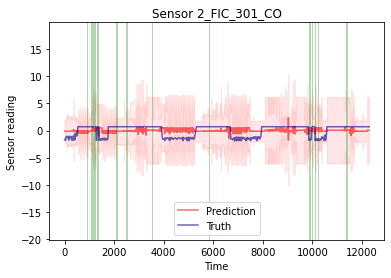} &
     \includegraphics[width=0.23\textwidth]{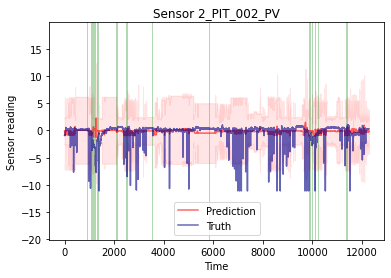}
	 \end{tabular}
	 \caption{Plot of examples of individual WADI sensor predictions (\textcolor{red}{red}, with  95\% confidence bounds)-vs-true values (\textcolor{blue}{blue}) over time. The highlighted \textcolor{green}{green} areas denote regions where the true anomalies occur in test data.}
    \label{fig:wadi_sensors}
\end{figure*}
\begin{figure}[!h]
    \centering
    \includegraphics[width=0.2\textwidth]{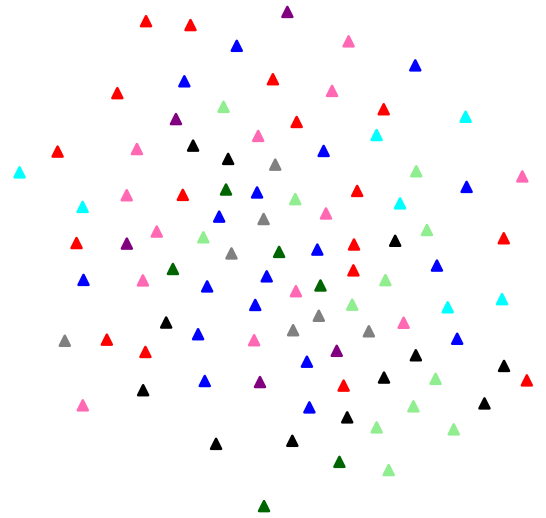}  \hspace{1.2cm}
    \includegraphics[width=0.2\textwidth]{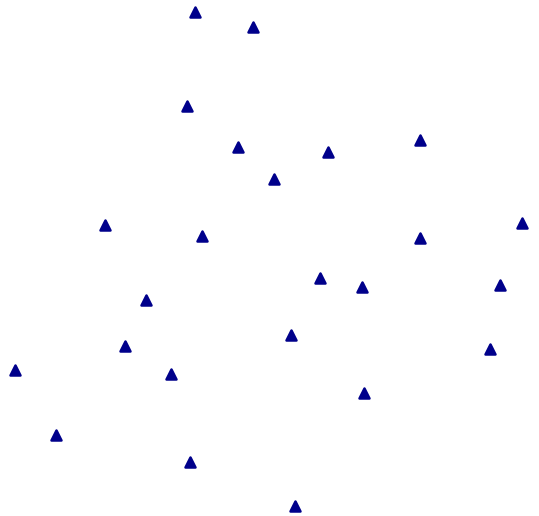}
    \caption{(Left) Plot of WADI Sensor Weights from GLUE using t-SNE \cite{van2014accelerating} illustrates that GLUE finds interesting relationships between the sensors and clusters similar sensors together (having the same colour). (Right) Plot of NASA Sensor Weights from GLUE using t-SNE \cite{van2014accelerating} illustrates that our model detects patterns and relationships between sensors.}
    \label{fig:nasa_text}
\end{figure}
We preprocessed the WADI and NASA datasets such that a single training example is window with $w = 5$ time steps. Table~\ref{tab:data} summarizes the statistics of our datasets \textit{i.e.} number of instances, sensors etc. The results of GLUE along with the other baselines (PCA, k-NN, AE, VAR, and GDN) are summarized in Tab.~\ref{tab:resultsWADI} and~\ref{tab:resultsNASA}. Both GDN and GLUE were trained for a total of 25 epochs on a quad-core Linux desktop (with hyper-threading enabled) with 16GB RAM. The models are trained using Adam optimizer with learning rate $1 \times 10^{-3}$ and $(\beta_1,\beta_2) = (0.9, 0.99)$. 
Through our experiments, we aim to answer the following research questions: 
\begin{enumerate}
    \item \textbf{RQ1.} How does GLUE compare with GDN and other baselines in terms of anomaly detection on the \texttt{NASA Turbofan} and \texttt{WADI} datasets? 
    \item \textbf{RQ2.} How does GLUE compare with GDN and VAR in terms of forecasting performance? 
    \item \textbf{RQ3.} Do the sensor embeddings learned by GLUE represent meaningful relationships? 
\end{enumerate}
Below, we discuss our results while answering the aforementioned research questions. Tables~\ref{tab:resultsWADI} and~\ref{tab:resultsNASA} show the anomaly detection results for the different models measured using \texttt{Precision}, \texttt{Recall} and \texttt{F1} metrics on the WADI and NASA datasets respectively. These tables demonstrate that GLUE performs at par with GDN on both the datasets while outperforming most baselines. On the \texttt{NASA Turbofan} dataset, however, K-NN beats both GLUE and GDN in terms of the \texttt{F1} score. This may be due to stationary nature of the dataset, which makes it extremely hard to distinguish between normal and anomalous data points based on forecasting error. 

Figure~\ref{fig:wadi_sensors} shows examples of individual WADI sensor true readings and mean of the forecasted predictions along with 95\% confidence intervals. Here the first and the fourth plots reveal circumstances when the true sensor readings deviate out of the predicted confidence bands contributing towards the max robust error (MRE) score of the system at that point in time. Second and third plots are examples of sensors which do not show anomalous signal (since the truth mostly lies within confidence bands of the predictions). 

Figure~\ref{fig:tuning} shows the Max Robust Errors computed over time on the WADI and NASA train-test datasets. The \textcolor{red}{red} points show the true anomalies in the respective datasets (none in WADI training data), whereas the \textcolor{blue}{blue} points represent the normal points for respective datasets. These plots help us visually see how GLUE separates out most of the anomalies and normal data.  
To answer our second research question, we compare the forecasting performance of VAR, GDN and GLUE in Tab.~\ref{tab:forecastingresults} for both the datasets. We can clearly see that on the \texttt{WADI} dataset, GLUE has the lowest mean squared error (MSE) while having a slightly high mean absolute error (MAE). VAR on the other hand has an orders of magnitude higher MSE. On the NASA dataset, GDN has the lowest MSE and MAE. 
\begin{table}[!htb]
  \centering
  \begin{tabular}{c|ccc}
    \Xhline{1pt}
     & \multicolumn{3}{c}{\texttt{WADI}} \\
     \textbf{Models} & \texttt{Precision(\%)} & \texttt{Recall(\%)} & \texttt{F1} \\
     \Xhline{1pt}
      \textbf{PCA} & 0.36 & 0.52 & 0.42 \\
     \textbf{K-NN} & 0.16 & 0.17 & 0.04 \\
     \textbf{Autoencoder} & 0.36 & 0.53 & 0.43 \\
     \textbf{VAR} & 0.50 & 0.50 & 0.50 \\
     \Xhline{1pt}
     \textbf{GDN} & \textbf{0.63} & \textbf{0.66} & \textbf{0.64} \\
     \textbf{GLUE} & 0.60 & 0.57 & 0.58 \\
     \Xhline{1pt}
  \end{tabular}
\vspace{1pt}
  \caption{Anomaly detection results obtained on WADI sensor datasets. GDN and GLUE perform significantly better than baselines and GLUE performs on par with GDN.}
  \label{tab:resultsWADI}
\end{table}
\begin{table}[!htb]
  \centering
  \begin{tabular}{c|ccc}
    \Xhline{1pt}
     & \multicolumn{3}{c}{\texttt{NASA Turbofan}} \\
     \textbf{Models} & \texttt{Precision(\%)} & \texttt{Recall(\%)} & \texttt{F1} \\
     \Xhline{1pt}
     \textbf{PCA} & 0.89 & 0.06 & 0.11\\
     \textbf{K-NN} & 0.55 & \textbf{0.55} & \textbf{0.55} \\
     \textbf{Autoencoder} & \textbf{0.97} & 0.14 & 0.25\\
     \textbf{VAR} & 0.50 & 0.50 & 0.50 \\
     \Xhline{1pt}
     \textbf{GDN} & 0.51 & 0.50 & 0.50 \\
     \textbf{GLUE} & 0.50 & 0.50 & 0.50 \\
     \Xhline{1pt}
  \end{tabular}
\vspace{1pt}
  \caption{Anomaly detection results obtained on NASA sensor datasets. NASA dataset is so stationary that the anomalies don't stand out. This could explain why baselines results do better, since they are simpler models.}
  \label{tab:resultsNASA}
\end{table}
\begin{table}[!htb]
  \centering
  \begin{tabular}{c|cc|cc}
    \Xhline{1pt}
     & \multicolumn{2}{c}{\texttt{WADI}} & \multicolumn{2}{c}{\texttt{NASA Turbofan}} \\
     \textbf{Models} & \texttt{MSE} & \texttt{MAE} & \texttt{MSE} & \texttt{MAE} \\
     \Xhline{1pt}
     \textbf{VAR} & 2497.52 & \textbf{0.65} & 129.92 & 0.99 \\
     \textbf{GDN} & 3.00 & \textbf{0.65} & \textbf{0.26} & \textbf{0.41} \\
     \textbf{GLUE} & \textbf{2.79} & 0.72 & 0.51 & 0.56 \\
     \Xhline{1pt}
  \end{tabular}
\vspace{1pt}
  \caption{Forecasting results obtained on two sensor datasets demonstrate that GLUE outperforms the competitive VAR baseline, but generally slight worse than GDN.}
  \label{tab:forecastingresults}
\end{table}
Figure \ref{fig:nasa_text} show TSNE plots of structural clusterings over sensors as learnt by GLUE models on WADI and NASA datasets respectively. WADI sensors have been colored by the available sensor grouping nomenclature. NASA dataset sensors are unlabeled.
\section{Conclusion}
\label{conclusion}
In this work, we propose GLUE (\textbf{G}DN with \textbf{L}ocal \textbf{U}ncertainty \textbf{E}stimation) building on the recently proposed Graph Deviation Network (GDN). GLUE not only automatically learns complex dependencies between variables and uses them to better identify anomalous behaviour, but also quantifies its predictive uncertainty, allowing us to account for the variation in the data as well to have more interpretable anomaly detection thresholds. Results on two real world datasets tell us that optimizing the negative Gaussian log likelihood is reasonable because GLUE’s forecasting results are at par with GDN and in fact better than the vector autoregressor baseline, which is significant given that at least GDN directly optimizes the MSE loss. In summary, our experiments demonstrate that GLUE is competitive with GDN at anomaly detection, with the added benefit of uncertainty estimations, as shown in Figure \ref{fig:wadi_sensors}. We also show that GLUE learns meaningful sensor embeddings which clusters similar sensors together. 

For future work, we aim to evaluate the quality of predictive uncertainties, which is a challenging task since there is no `ground truth' uncertainty estimate, in lines similar to a recent work by Lakshminarayanan et al \cite{lakshminarayanan2016simple}. We also aim to increase the number of evaluation datasets and to look into more precise uncertainty estimates. 

\bibliographystyle{IEEEtran}
\bibliography{bibliography}




\end{document}